\documentclass[conference]{IEEEtran}
\usepackage{times}
\usepackage[svgnames]{xcolor}

\usepackage{amssymb}
\usepackage{amsmath} 
\usepackage{inconsolata}
\usepackage{graphicx}
\usepackage{float}
\usepackage{subfigure}
\usepackage{multirow}
\usepackage{setspace}

\usepackage{soul}
\usepackage{color}
\soulregister{\cite}7 
\soulregister{\citep}7 
\soulregister{\citet}7 
\soulregister{\ref}7 
\soulregister{\pageref}7 
\usepackage{booktabs,xcolor,siunitx}

\usepackage[linesnumbered,ruled,vlined]{algorithm2e}

\usepackage[numbers]{natbib}
\usepackage{multicol}

\usepackage[bookmarks=true]{hyperref}
\hypersetup{hidelinks,
            colorlinks=true,
            linkcolor=LightSlateGrey,
            urlcolor=LightSlateGrey,
            citecolor=LightSlateGrey}

\newcommand{\etal}[0]{\textit{et al.}}
\newcommand{\eg}[0]{\textit{e.g.}}

\newcommand{\ie}[0]{\textit{i.e.}}

\begin{document}

\title{VLMPC: Vision-Language Model Predictive Control for Robotic Manipulation}





%

\author{\authorblockN{Wentao Zhao\footnotemark$^\text{*}$,
Jiaming Chen\footnotemark$^\text{*}$,
Ziyu Meng, 
Donghui Mao, 
Ran Song\footnotemark$^\text{\dag}$, Wei Zhang}
\authorblockA{School of Control Science and Engineering, Shandong University, China\\
E-mail: \textit{zwt7@mail.sdu.edu.cn; ppjmchen@gmail.com; mziyu@mail.sdu.edu.cn; } \\ \textit{donghui.mao@mail.sdu.edu.cn; ransong@sdu.edu.cn; davidzhang@sdu.edu.cn}
}}

\maketitle

\renewcommand{\thefootnote}{\fnsymbol{footnote}}
\footnotetext[1]{These authors contributed equally to this work.}
\footnotetext[2]{Corresponding author: Ran Song.}

\begin{abstract}
 Although Model Predictive Control (MPC) can effectively predict the future states of a system and thus is widely used in robotic manipulation tasks, it does not have the capability of environmental perception, leading to the failure in some complex scenarios. To address this issue, we introduce \textbf{V}ision-\textbf{L}anguage \textbf{M}odel \textbf{P}redictive \textbf{C}ontrol (\textbf{VLMPC}), a robotic manipulation framework which takes advantage of the powerful perception capability of vision language model (VLM) and integrates it with MPC. Specifically, we propose a conditional action sampling module which takes as input a goal image or a language instruction and leverages VLM to sample a set of candidate action sequences. Then, a lightweight action-conditioned video prediction model is designed to generate a set of future frames conditioned on the candidate action sequences. VLMPC produces the optimal action sequence with the assistance of VLM through a hierarchical cost function that formulates both pixel-level and knowledge-level consistence between the current observation and the goal image. We demonstrate that VLMPC outperforms the state-of-the-art methods on public benchmarks. More importantly, our method showcases excellent performance in various real-world tasks of robotic manipulation. Code is available at~\url{https://github.com/PPjmchen/VLMPC}.
\end{abstract}

\IEEEpeerreviewmaketitle

\section{Introduction}

Burgeoning foundation models~\cite{gpt4,gpt3, palm,bommasani2021opportunities,palm-e} have demonstrated powerful capabilities of knowledge extraction and reasoning. Exploration based on foundation models has thus flourished in many fields such as computer vision~\cite{liu2023visual,chen2023minigpt,dai2023instructblip,bai2023sequential}, AI for science~\cite{bi2023accurate}, healthcare~\cite{moor2023foundation,thirunavukarasu2023large,zhou2023foundation,qiu2023large}, and robotics~\cite{rt-2,ha2023scaling,ren2023robots,yu2023language,mandi2023roco}. Recently, a wealth of work has made significant progress in incorporating foundation models into robotics. These works usually leveraged the strong understanding and reasoning capabilities of versatile foundation models on multimodal data including language~\cite{voxposer,rt-2,ren2023robots,yu2023language,languagempc,mandi2023roco}, image~\cite{voxposer,liu2023aligning} and video~\cite{rt-2} for enhancing robotic perception and decision-making. 

To achieve knowledge transfer from foundation models to robots, most early works concentrate on robotic planning~\cite{huang2022language,huang2022inner,chen2023open,wang2023voyager,singh2023progprompt,raman2022planning,song2023llm,liu2023llm+,lin2023text2motion,ding2023task,yuan2023plan4mc,xie2023translating,lu2023multimodal,pallagani2024prospects,ni2023grid}, which directly utilize large language models (LLMs) to decompose high-level natural language command and abstract tasks into low-level and pre-defined primitives (\textit{i.e.,} executable actions or skills). Although such schemes intuitively enable robots to perform complex and long-horizon tasks, they lack the capability of visual perception. Consequently, they heavily rely on pre-defined individual skills to interact with specific physical entities, which limits the flexibility of robotic planning. Recent works~\cite{voxposer,rt-2,wake2023gpt,hu2023look} remedy this issue by integrating with large-scale vision-language models (VLMs) to improve scene perception and generate trajectories adaptively for robotic manipulation in intricate scenarios without using pre-defined primitives. 

Although existing methods have shown promising results in incorporating foundation models into robotic manipulation, interaction with a wide variety of objects and humans in the real world remains a challenge. Specifically, since the future states of a robot are not fully considered in the decision-making loop of such methods, the reasoning of foundation models is primarily based on current observations, resulting in insufficient forward-looking planning. For example, when opening a drawer, the latest method based on VLM~\cite{voxposer} cannot directly generate an accurate trajectory to pull the drawer handle due to the lack of prediction on the future state, and thus it still requires to design specific primitives on object-level interaction. Hence, it is desirable to develop a robotic framework that performs with a human-like \emph{``look before you leap''} ability.

Model predictive control (MPC) is a control strategy widely used in robotics~\cite{shim2003decentralized,allibert2010predictive,howard2010receding,williams2017information,lenz2015deepmpc}. MPC possesses an appealing attribute of predicting the future states of a system through a predictive model. Such forward-looking attribute allows robots to plan their actions by considering potential future scenarios, thereby enhancing their ability to interact dynamically with various environments. Traditional MPC~\cite{shim2003decentralized,howard2010receding,williams2017information,mpc_robot_exp2,mpc_robot_exp3} usually builds a deterministic and sophisticated dynamic model corresponding to the task and environment, which does not adapt well to intricate scenes in the real world.  Recent research \cite{visual_foresight,ye2020object,nair2022learning,3dmpc,vp2,ebert2018robustness} has explored using vision-based predictive models to learn dynamic models from visual inputs and predict high-dimensional future states in 2D~\cite{visual_foresight,ye2020object,vp2,ebert2018robustness} or 3D~\cite{visual_foresight,nair2022learning,3dmpc,ebert2018robustness} spaces. Such methods are based on current visual observation for proposing manipulation actions in the MPC loop, which enables robots to make more reasonable decisions based on visual clues. Nevertheless, the effectiveness of such methods is constrained by the limitations inherent in visual predictive models trained on finite datasets. Such models struggle to accurately predict scenarios involving scenes or objects they have not previously encountered. This issue becomes especially pronounced in the real-world environments often partially or even fully unseen to robots, where the models can only perform basic tasks that align closely with their training data.


Naturally, large-scale vision-language models have the potential to address this problem by providing extensive open-domain knowledge and offering a more comprehensive understanding of diverse and unseen scenarios, thereby enhancing the predictive accuracy and adaptability of the scheme for robotic manipulation. Thus, this work presents the \textbf{V}ision-\textbf{L}anguage \textbf{M}odel \textbf{P}redictive \textbf{C}ontrol (\textbf{VLMPC}), a framework that combines VLMs and model predictive control to guide the robotic manipulation with complicated path planning including rotation and interaction with scene objects. By leveraging the strong ability of visual reasoning and visual grounding for sampling actions provided by VLM, VLMPC avoids the manual design of individual primitives, and addresses the limitation that previous methods based on VLMs can only compose coarse trajectories without foresight.

As shown in Fig.~\ref{fig1}, VLMPC takes as input either a goal image indicating the prospective state or a language instruction. We propose an action sampling module that uses VLM to initialize the task and handle the current observation, which generates a conditional action sampling distribution for further producing a set of action sequences. With the action sequences and the history image observation, VLMPC adopts a lightweight action-conditioned video prediction model to predict a set of future frames. To assess the quality of the candidate action sequences through the future frames, we also design a hierarchical cost function composed of two sub-costs: a pixel-level cost measuring the difference between the video predictions and the goal image and a knowledge-level cost making a comprehensive evaluation on the video predictions. VLMPC finally chooses the action sequence corresponding to the best video prediction, and then picks the first action from the sequence 
to execute while feeding the subsequent actions into the action sampling module combined with conditional action sampling.

The main contributions of this paper are as follows:
\begin{itemize}
    \item We propose VLMPC for robotic manipulation, which incorporates a learning-based dynamic model to predict future video frames and seamlessly integrates VLM into the MPC loop for open-set knowledge reasoning. 
    
    \item We design a conditional action sampling module to sample robot actions from a visual perspective and a hierarchical cost function to provide a comprehensive and coarse-to-fine assessment of video predictions.
    
    \item We conduct experiments in both simulated and real-world scenes to demonstrate that VLMPC provides good knowledge reasoning and effective foresight, achieving state-of-the-art performance without any primitives.
\end{itemize}

\section{Related Work}
Since the proposed VLMPC integrates MPC with foundation models, this section reviews them in the context of robotic manipulation. 

\subsection{Model Predictive Control for Robotic Manipulation}
Model predictive control (MPC) is a multivariate control algorithm widely used in robotics \cite{shim2003decentralized,allibert2010predictive,howard2010receding,williams2017information,lenz2015deepmpc,mpc-policy,mpc_robot_exp1,mpc_robot_exp2,mpc_robot_exp3,voxposer,visual_foresight}. It employs a predictive model to estimate future system states, subsequently formulating the control law through solving a constrained optimization problem~\cite{mpc_review,mpc-policy}. The foresight capability of MPC, combined with its constraint-handling features, enables the development of advanced robotic systems which operate safely and efficiently in variable environments~\cite{howard2010receding}.

In the context of robotic manipulation, the role of MPC is to make the robot manipulator move and act in an optimal way with respect to input and output constraints~\cite{bhardwaj2022storm,visual_foresight,deep_visual_foresight,3dmpc,ye2020object,nair2022learning,vp2}.
In particular, action-based predictive models are frequently used in MPC for robotic manipulation, referring to a prediction model designed to forecast the potential future outcomes of specific actions, connecting sequence data to decision-making processes.
Bhardwaj \etal~\cite{bhardwaj2022storm} proposed a sampling-based MPC integrated with low discrepancy sampling, smooth trajectory generation, and behavior-based cost functions to produce good robotic actions reaching the goal poses.
Visual Foresight~\cite{visual_foresight,deep_visual_foresight} first generated robotic planning towards a specific goal by leveraging a video prediction model to simulate candidate action sequences and then scored them based on the similarity between their predicted futures and the goal. Xu \etal~\cite{3dmpc} proposed a 3D volumetric scene representation that simultaneously discovers, tracks, and reconstructs objects and predicts their motion under the interactions of a robot. Ye \etal~\cite{ye2020object} presented an approach to learn an object-centric forward model, which planned for action sequences to achieve distant desired goals. Recently, Tian \etal~\cite{vp2} conducted a simulated benchmark for action-conditioned video prediction in the form of MPC framework that evaluated a given model for simulated robotic manipulation through sampling-based planning.

Recently, some video prediction models independent of the MPC framework have also been proposed for robotic manipulation. For instance, VLP~\cite{du2024video} and UniPi~\cite{du2023learning} combined text-to-video models with VLM to generate long-horizon videos used for extracting control actions. V-JEPA~\cite{bardes2024vjepa} developed a latent video prediction strategy to make predictions in a learned latent space. Similarly, Dreamer~\cite{Hafner2020Dream} learned long-horizon behaviors through predicting state values and actions in a compact latent space where the latent states have a small memory footprint. RIG~\cite{nair2018visual} used a latent variable model to generate goals for the robot to learn diverse behaviors. Planning~to~Practice~\cite{fang2022planning} proposed a sub-goal generator to decompose a goal-reaching task hierarchically in the latent space.



\subsection{Foundation Models for Robotic Manipulation}

Foundation models are large-scale neural networks trained on massive and diverse datasets~\cite{bommasani2021opportunities}. Breakthroughs such as GPT-4, Llama and PaLM exemplify the scaling up of LLMs~\cite{gpt4,gpt3,llama,palm}, showcasing notable progress in knowledge extraction and reasoning. Simultaneously, there has been an increase in the development of large-scale VLMs~\cite{flamingo,clip,align,dalle,palm-e,qwen}. VLMs typically employ cross-modal connectors to merge visual and textual embeddings into a unified representation space, enabling them to process multimodal data effectively. 

The application of foundation models in advanced robotic systems is an emerging research field. Many studies focus on employing LLMs for knowledge reasoning and robotic manipulation~\cite{huang2022inner,zeng2022socratic,huang2023grounded,liang2023code,hu2023look}. To allow LLMs to perceive the physical environments, auxiliary modules such as textual descriptions of the scene~\cite{huang2022inner,zeng2022socratic}, affordance models~\cite{huang2023grounded}, and perception APIs~\cite{liang2023code} are essential. Additionally, using VLMs for robotic manipulation is being explored~\cite{voxposer,palm-e,rt-2}. For example, PaLM-E enhanced the understanding of robots with regard to complex visual-textual tasks~\cite{palm-e}, while RT-2 specialized in real-time image processing and decision-making~\cite{rt-2}. However, most existing methods are limited by their reliance on pre-defined executable skills or hand-designed motion primitives~\cite{liang2023code,voxposer}, constraining the adaptability of robots in complex, real-world environments and their interaction with diverse, unforeseen objects.

\textbf{Difference from closely related works.} This work is closely related to some MPC-based methods~\cite{visual_foresight,deep_visual_foresight,vp2,3dmpc} designed for robotic manipulation. However, most of these methods were designed for manipulation tasks merely involving specific objects as regular MPC has limitations in two aspects: (1) The predictive models used in regular MPC are constrained with small-scale training datasets, and thus cannot precisely predict the process of interaction with objects unseen during training; (2) The cost functions of regular MPC are usually designed with a defined set of constraints such as physical limitations or operational safety margins. Although these constraints ensure that robot actions adhere to them while striving for optimal performance, accurately modeling such constraints is highly difficult in real-world scenarios. To address the above two problems, the proposed VLMPC leverages a video prediction model which is trained with a large-scale robot manipulation dataset~\cite{padalkar2023open} and can be directly transferred to the real world. Also, VLMPC incorporates powerful VLMs into cost functions with high-level knowledge reasoning, which provides constraints produced through interactions with open-set objects.

\begin{figure*}[t]
    \centering
    \includegraphics[height=9.3cm]{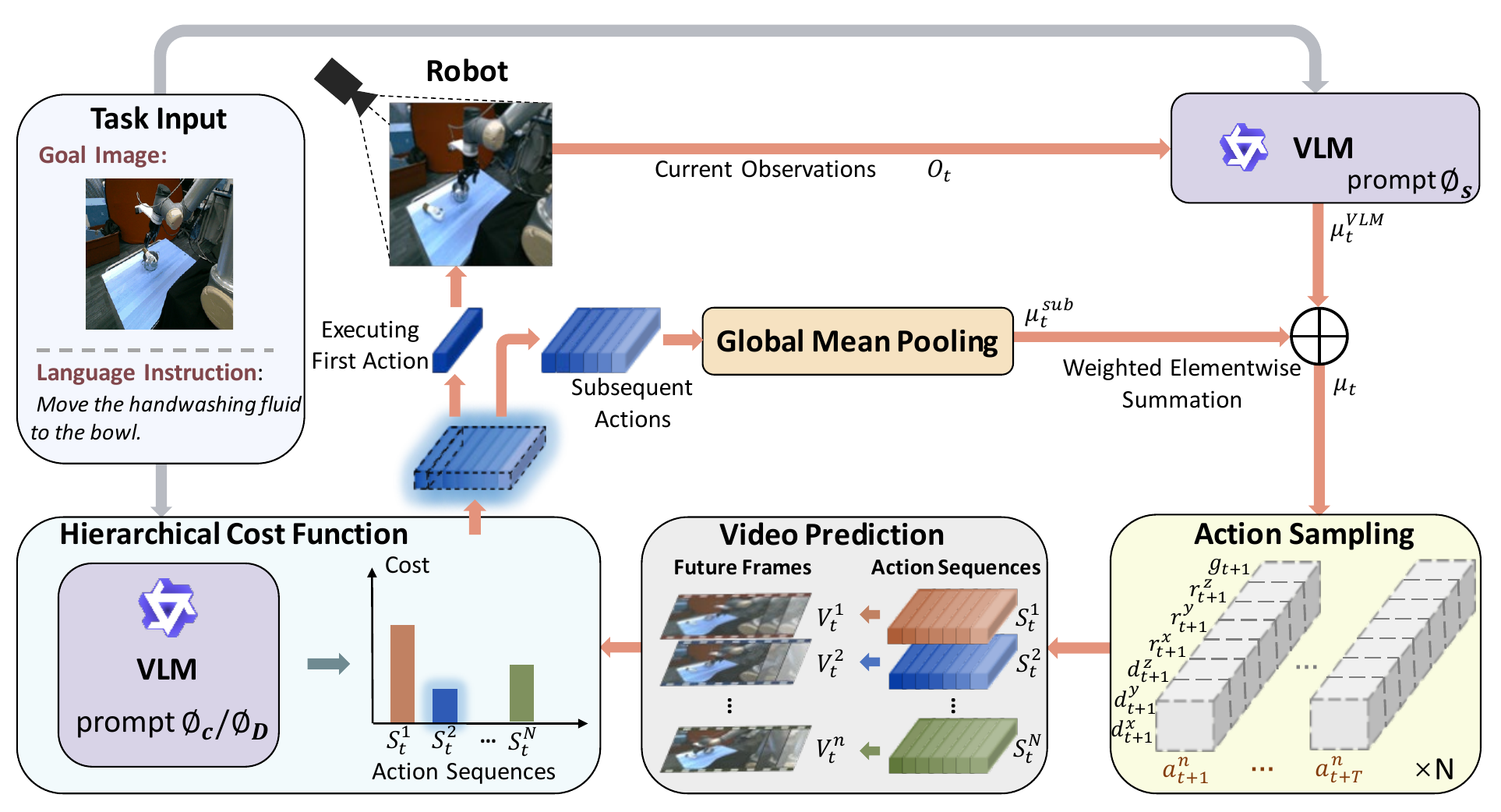}
    \caption{VLMPC takes as input either a goal image or a language instruction. It first prompts VLMs to generate a conditional sampling distribution, from which action sequences are derived. Then, such action sequences are fed into a lightweight action-conditioned video prediction model to predict a set of future frames. The assessment of VLMPC is performed with a hierarchical cost function composed of two sub-costs: a pixel distance cost and a VLM-assisted cost for performing video assessments based on the future frames. VLMPC finally selects the best action sequence, in which the robot picks the first action to execute and the subsequent actions are fed into the action sampling module to further assist conditional action sampling.}.
    \label{fig1}
\end{figure*}

\section{Method}
Fig.~\ref{fig1} illustrates the overview of the VLMPC framework. It takes as input either a goal image indicating the prospective state or a language instruction that depicts the required manipulation, and performs a dynamic strategy that iteratively makes decisions based on the predictions of future frames. First, a conditional action sampling scheme is designed to prompt VLMs to take into account both the input and the current observation and reason out prospective future movements, from which a set of candidate action sequences are sampled. Then, an action-conditioned video prediction model is devised to predict a set of future frames corresponding to the sampled action sequences. Finally, a hierarchical cost function including two sub-costs and a VLM switcher are proposed to comprehensively compute the coarse-to-fine scores for the video predictions and select the best action sequence. The first action in the sequence is fed into the robot for execution, and the subsequent actions go through a weighted elementwise summation with the conditional action distribution. We elaborate each component of VLMPC in the following.

\subsection{Conditional Action Sampling}
In an MPC framework, $N$ candidate action sequences $\mathcal{S}_t = \{S^1_t, S^2_t, ..., S^N_t\}$ are sampled from a custom sampling distribution at each step $t$, where $S^n_t = \{a^{n}_{t+1}, a^{n}_{t+2}, ..., a^{n}_{t+T}\}$ contains $T$ actions and $n \in \{1,...,N\}$. For every $\tau \in \{t+1,...,t+T\}$ representing a future step after $t$, $a^{n}_{\tau} \in \mathbb{R}^7$ is a $7$-dimensional vector composed of the movement $[d^x_\tau, d^y_\tau, d^z_\tau]$ of the end-effector in Cartesian space, the rotation $[r^x_\tau, r^y_\tau, r^z_\tau]$ of the gripper, and a binary grasping state $g_t$ indicating the open or close state of the end-effector. 

\begin{figure}[t]
    \centering
    \includegraphics[width=\linewidth]{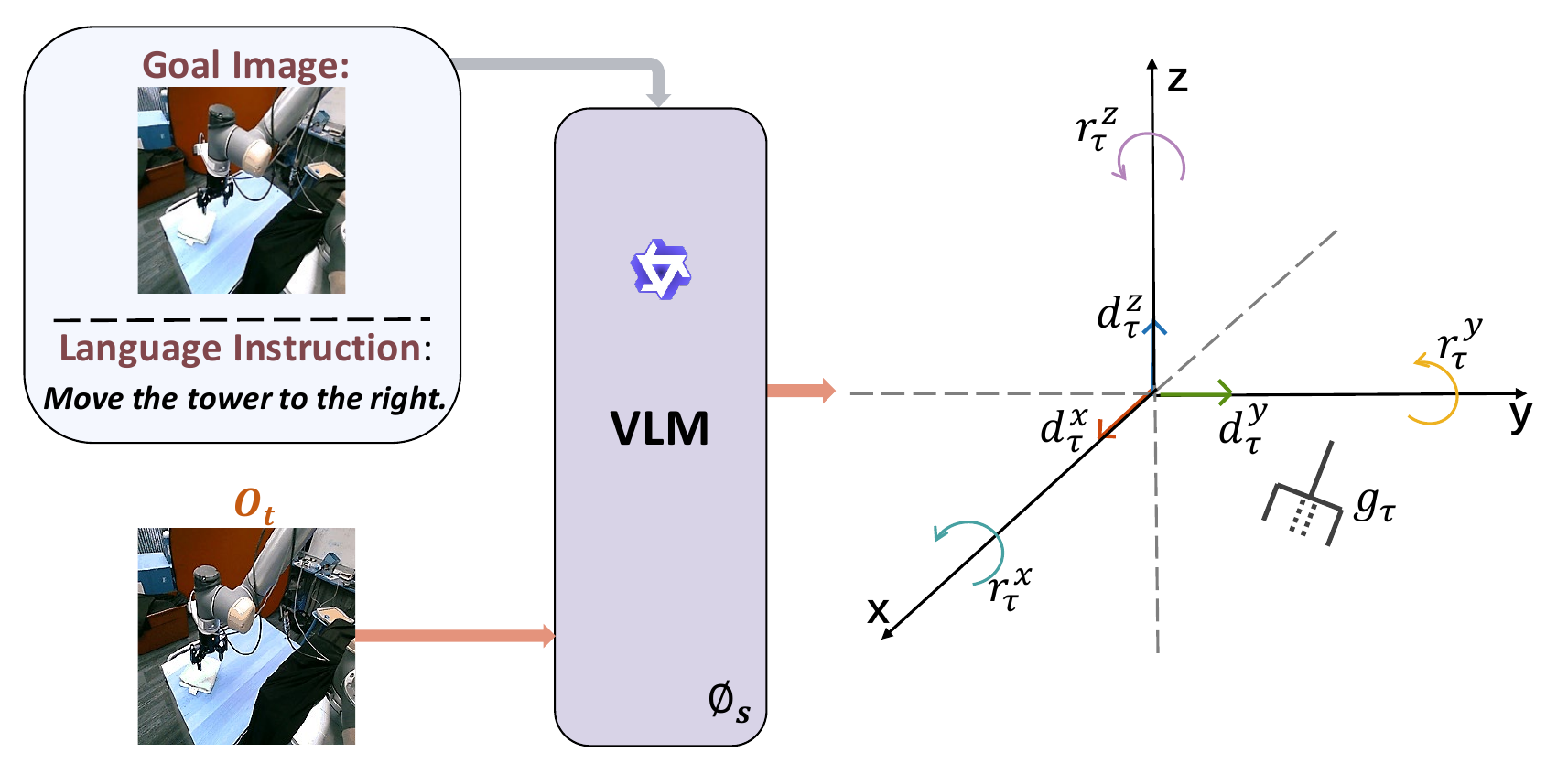}
    \caption{The VLMs subject to a specifically designed prompt $\phi_s$ take as input the current observation $O_t$ and a goal image or a language instruction to generate an end-effector moving direction at coarse level.} 
    \label{fig2}
\end{figure}

Given a goal image $G$ or a language instruction $L$ as the input of VLMPC along with the current observation $O_t$, we expect VLMs to generate appropriate future movements, from which a sampling distribution is derived for action sampling. As shown in Fig.~\ref{fig2}, the current observation $O_t \in \mathbb{R}^{h\times w \times 3}$ is represented as an RGB image with the shape of $h\times w \times 3$ taken by an external monocular camera. We design a prompt $\phi_s$ that drives VLMs to analyze $O_t$ alongside the input. $\phi_s$ forces VLMs to identify and localize the object with which the robot is to interact, reason about the manner of interaction, and generate appropriate future movements. The output of VLMs can be formulated as
\begin{equation}
\label{eq1}
    \text{VLM}(O_t, G  \lor L|\phi_s) = \{\widehat{d}^x_t, \widehat{d}^y_t, \widehat{d^z_t}, \widehat{r^x_t}, \widehat{r^y_t}, \widehat{r^z_t}, g_t\} 
\end{equation}
where $\widehat{\cdot} \in \{+1, 0, -1\}$ denotes the predicted moving/rotation direction alongside the corresponding axis and $g_t \in \{0, 1\}$ represents the predicted binary state of the end-effector.

To obtain a set of candidate action sequences, we follow the scheme of Visual Foresight~\cite{visual_foresight} and adopt Gaussian sampling that samples $N$ action sequences with the expected movement in each action dimension as the mean. Hence we further map the output of VLMs into a sampling mean $\mu^\text{VLM}_t$:
\begin{equation}
\label{eq2}
    \mu^\text{VLM}_t = w_m * \{\widehat{d}^x_t, \widehat{d}^y_t, \widehat{d}^z_t\} \cup w_r * \{\widehat{r}^x_t, \widehat{r}^y_t, \widehat{r}^z_t\} \cup \{g_t\}
\end{equation}
where $w_m$ and $w_r$ are hyperparameters for mapping the output of VLMs into the action space of the robot.

Hallucination phenomenon is a common issue which hinders the stable use of large-scale VLMs in real-world deployment, as it may
result in unexpected consequences caused by incorrect understandings of the external environment. To mitigate the hallucination phenomenon, we propose to make use of the historical information derived from the subsequent candidate action sequence of the last step. This leads to another sampling mean $\mu^\text{sub}_t$. Please refer Sec.~\ref{cost} for the detailed process of obtaining $\mu^\text{sub}_t$. Then we perform a weighted elementwise summation of $\mu^\text{sub}_t$ and $\mu^\text{VLM}_t$ to produce the final sampling mean $\mu_t$ of step $t$:
\begin{equation}
\label{eq3}
    \mu_t =w_\text{VLM} * \mu^\text{VLM}_t + w_\text{sub} * \mu^\text{sub}_t
\end{equation}
where $w_\text{VLM}$ and $w_\text{sub}$ are weighting parameters. Finally, we sample $S_t$ from the Gaussian distribution $S_t^n\sim\mathcal{N}(\mu_t, I)$ repeatedly $N$ times.

This conditional action sampling scheme allows VLMs to provide the guidance of robotic manipulation at a coarse level via knowledge reasoning from the image observation and the task goal. Next, with the candidate action sequences, we introduce the module for action-conditioned video prediction.

\subsection{Action-Conditioned Video Prediction}

Given the candidate action sequences, it is necessary to estimate the future state of the system when executing each sequence, which provides the forward-looking capability of VLMPC.

Traditional MPC methods often rely on hand-crafted deterministic dynamic models. Developing and refining such models typically require extensive domain knowledge, and they may not capture all relevant dynamics. On the contrary, video is rich in semantic information and thus enables the model to handle complex, dynamic environments more effectively and flexibly. Moreover, video can be directly fed into a VLM for knowledge reasoning. Thus, we use the action-conditioned video prediction model to predict the future frames corresponding to candidate action sequences. 

\begin{figure}[t]
    \centering
    \includegraphics[width=\linewidth]{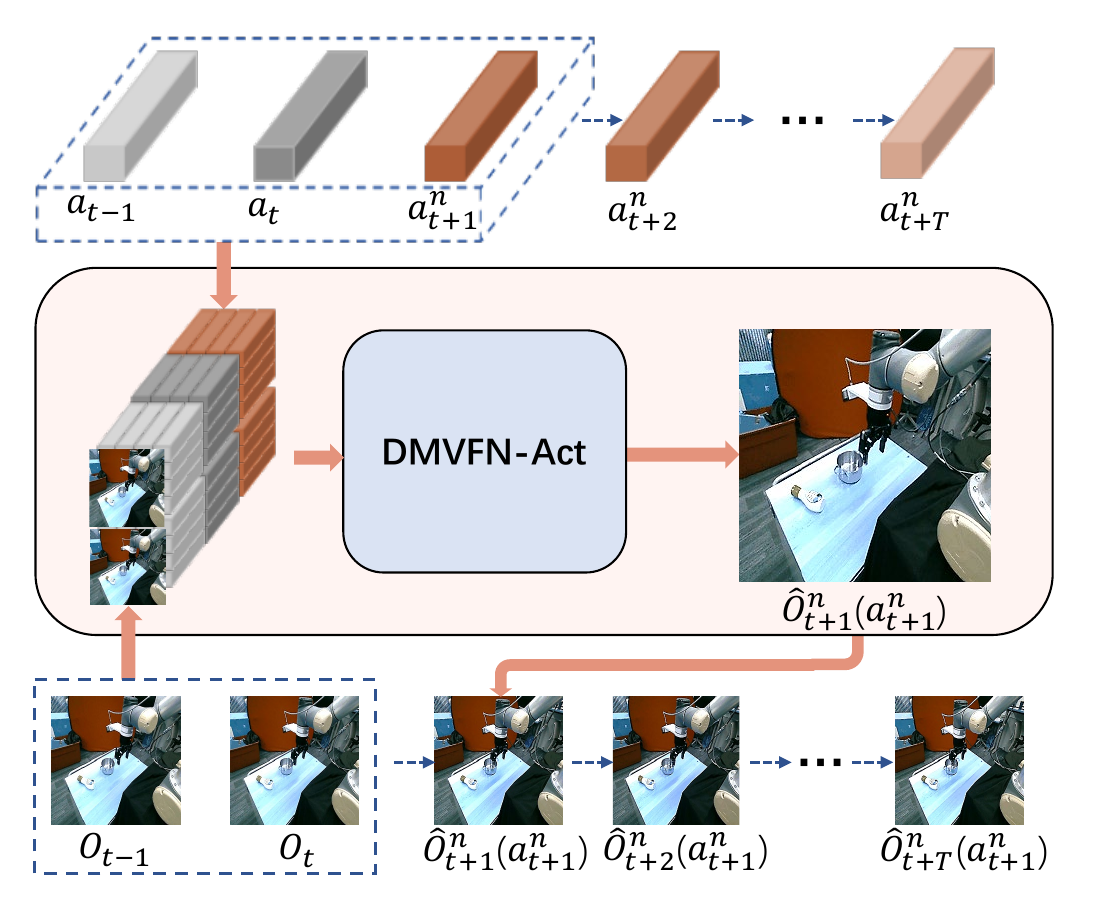}
    \caption{Given the past two frames $O_{t}$ and $O_{t-1}$ with the executed actions $a_{t-1}$ and $a_{t}$ corresponding to them and the action $a^{n}_{t+1}$,  DMVFN-Act predicts the next frame $\widehat{O}^{n}_{t+1}(a^{n}_{t+1})$. The dashed boxes and arrows indicate the iterative process of taking the actions one by one and predicting the future states frame by frame.}
    \label{fig3}
\end{figure}

We build a variant version of DMVFN~\cite{dmvfn}, an efficient dynamic multi-scale voxel flow network for video prediction, to perform action-conditioned video prediction. We name it DMVFN-Act. Given the past two historical frames $O_{t-1}$ and $O_t$, DMVFN predicts a future frame $\widehat{O}_{t+1}$, formulated as
\begin{equation}
\label{eq4}
    \widehat{O}_{t+1} = \text{DMVFN}(O_{t-1}, O_{t})
\end{equation}

With the candidate action sequences $S_t$ and the corresponding executed actions $a_{t-1}$ and $a_t$, we expect DMVFN-Act to take the actions one by one and predict future states frame by frame as illustrated in Fig.~\ref{fig3}. For simplicity, we explain this process by taking one sequence $S^n_t = \{a^{n}_{t+1}, a^{n}_{t+2}, ..., a^{n}_{t+T}\}$ as example. We broadcast $a_{t-1}$, $a_t$, $a^{n}_{t+1} \in \mathbb{R}^7$ to the image size $a_{t-1}'$, $a_t'$, ${a^{n}_{t+1}}' \in \mathbb{R}^{h\times w\times 7}$, and then concatenate them 
with  $O_{t-1}$ and $O_t$ respectively, formulated as
\begin{equation}
\begin{split}
    O_{t-1}' &= [O_{t-1}\cdot a_{t-1}' \cdot a_t' \cdot {a^{n}_{t+1}}'], \\
    O_{t}' &= [O_{t}\cdot a_{t-1}' \cdot a_t' \cdot {a^{n}_{t+1}}']
     \label{eq5}
\end{split}
\end{equation}
where $[\cdot]$ represents the channelwise concatenation, and $O_{t-1}'$ and $O_{t}'$ denote the action-conditioned historical observations. In DMVFN-Act, the input layer is modified to adapt $O_{t-1}'$ and $O_{t}'$ and predict one future frame $\widehat{O}^{n}_{t+1}(a^{n}_{t+1})$ conditioned by the candidate action $a^{n}_{t+1}$, expressed as
\begin{equation}
    \widehat{O}^{n}_{t+1}(a^{n}_{t+1}) = \text{DMVFN-Act}(O_{t-1}', O_{t}')
    \label{eq6}
\end{equation}
DMVFN-Act iteratively predicts future frames via Eqs.~(\ref{eq5}) and  (\ref{eq6}) until all candidate actions are used . The action-conditioned video prediction can be represented as:
\begin{equation}
\label{eq7}
    V^n_t = \{
    \widehat{O}^{n}_{t+1}(a^{n}_{t+1}), \widehat{O}^{n}_{t+2}(a^{n}_{t+2}),..., \widehat{O}^{n}_{t+T}(a^{n}_{t+T})\}
\end{equation}

To improve efficiency, the $N$ candidate action sequences are organized into a batch and predict all the action-conditioned videos $V_t=\{V^1_t, V^2_t, ..., V^N_t\}$ at step $t$ in one inference.

\subsection{Hierarchical Cost Function}\label{cost}
To comprehensively assess the video predictions, we design a cost function composed of two sub-costs that provide a hierarchical assessment at pixel and knowledge levels, respectively. We also propose a VLM Switcher which dynamically selects one or both sub-costs in a manner adaptive to the observation.

\subsubsection{Pixel Distance Cost}
While the task input is the goal image $G$, an intuitive way to assess video predictions is to sum the pixel distances between each future frame and the goal image. Following Visual Foresight~\cite{visual_foresight}, we calculate the $l_2$ distance between each future frame $\widehat{O}^{n}_{\tau}(a^{n}_{\tau})$ in an action-conditioned video $V_t^n$ and $G$, and then sum the distances as the pixel distance cost $C_P^n(t)$ for $V_t^n$ over $\tau$:
\begin{equation}
\label{eq8}
    C_P^n(t)=\sum\limits_{\tau=t+1}^{t+T}||\widehat{O}^{n}_{\tau}(a^{n}_{\tau})-G||_2
\end{equation}
Then, the pixel distance cost $C_P(t)$ at step $t$ for $V_t$ can be computed as
\begin{equation}
\label{eq9}
    C_P(t)=\{C_P^n(t) |n\in \{1,2,..., N\}\}
\end{equation}

\begin{figure}[t]
    \centering
    \includegraphics[width=\linewidth]{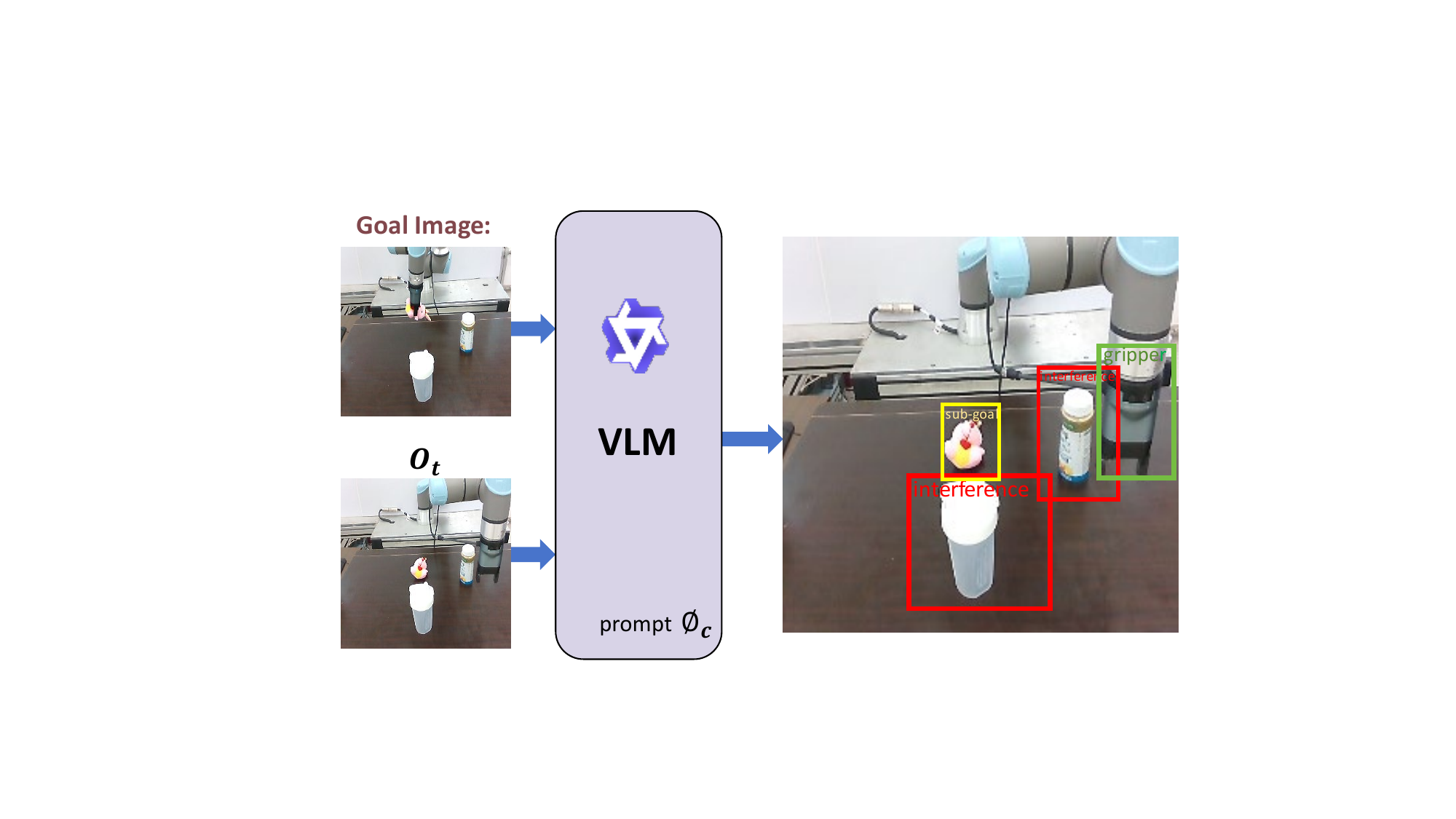}
    \caption{Illustration of the end-effector, the next sub-goal and the
interference objects in the current observation. Red, green, and yellow boxes denote the interference objects, the end-effector and the next sub-goal generated by VLMPC.}
    \label{fig4}
\end{figure}

The pixel distance cost encourages the robot to move directly towards the goal position in accordance with the goal image. This cost is simple yet effective when the task contains only one sub-goal, \eg, \textit{push a button}. However, for tasks that require manipulating objects with multiple sub-goals, where a common type is \textit{taking an object from position A to B}, this cost usually guide the robot to move directly towards \textit{position B} to reduce the pixel distance. To facilitate such situations, we further introduce the VLM-assisted cost. 

\subsubsection{VLM-Assisted Cost}
Many robotic manipulation tasks contain multiple sub-goals and interference objects, which require knowledge-level task planning. For example, in the task of `\textit{grasp the bottle and put it in the bowl, while watching out the cup}', the bottle should be identified as the sub-goal before the robot grasps it, and the bowl is the next sub-goal after the bottle is grasped, and the cup is an interference object. 
It is thus critical to dynamically identify the sub-goals and interference objects in each step, and make appropriate assessments on the action-conditioned video predictions so that we can select the best candidate action sequence to achieve the sub-goals as long as avoiding the interference object. Moreover,  We design a VLM-assisted cost to realize it at the knowledge level.

\begin{algorithm}[t]\label{algo}
    \SetAlgoNoLine
    \caption{VLMPC}
    \KwIn{Goal image $G$ or language instruction $L$, and obvevation $O_t$ at every step}
    \BlankLine
    \While{task not done $\text{\textbf{or}}$ $t\leq T_{max}$}{
        Generates a sampling distribution by VLM
        $D(\mu^{\text{VLM}}) \leftarrow \text{VLM}(O_t, G  \lor L|\phi_s)$\;
        Refine it with historical information $\mu^\text{sub}_t$ \
        $D(\mu_t) = D(w_\text{VLM} * \mu^\text{VLM}_t + w_\text{sub} * \mu^\text{sub}_t$)\;
        $\mathcal{S}_t \leftarrow$ sample $N$ action sequences\;
        \ForEach{sequence $S^n_t \in \mathcal{S}_t$}{
            \For{future step $\tau=t+1,...,t+T$}{
                $\widehat{O}^{n}_{\tau}(a^{n}_{\tau}) \leftarrow$ predict the future frame\;
            }
            $V^n_t=\{\widehat{O}^{n}_{\tau}(a^{n}_{\tau})|\tau \in \{t+1,...,t+T\}$ \;
        }
        $C_P(t) \leftarrow$ calculate the pixel distance cost\;
        $C_\text{VLM} \leftarrow$ calculate the VLM-assisted cost\;
        $C_t{ \leftarrow}$ arrange cost through VLM swicher\;
        $S^{n^{\star}}_t \leftarrow$ select the optimal action sequence\;
        Execute the first action $a^{n^{\star}}_{t+1}$ in the optimal sequence\;
        Update $\mu_{t+1}^{sub}$ using $\{a^{n^{\star}}_{\tau}|\tau \in \{t+2,...,t+T\}\}$\;
    }
\end{algorithm}

Specifically, with the current observation $O_t$ and the task input $G$ or $L$, we design a prompt $\phi_C$ to drive VLMs to reason out and localize the next sub-goal and all the interference objects, where the sub-goal is usually the next object to interact with the robot. As illustrated in Fig.~\ref{fig4}, this process yields the bounding boxes of the robot's end-effector $e_t$, the next sub-goal $s_t$ and all the interference objects $I_t$ in the current observation:
\begin{equation}
\label{eq10}
    \text{VLM}(O_t, G \lor L | \phi_C) = \{e_t, s_t, I_t\}
\end{equation}

\begin{figure*}[t]
    \centering
    \includegraphics[height=5.5cm]{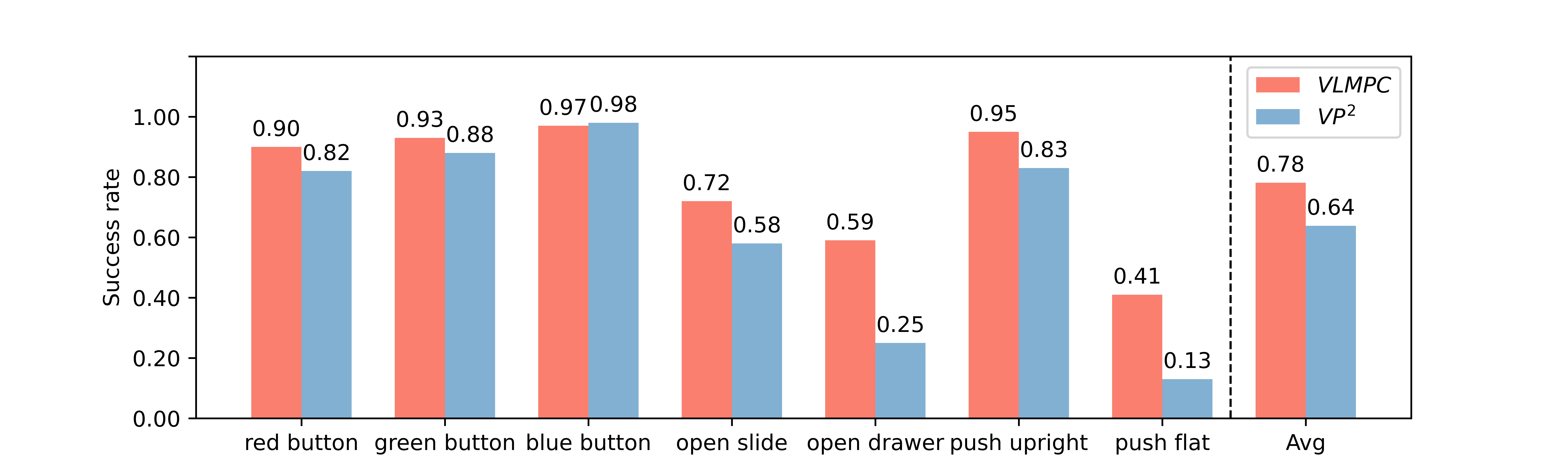}
    \caption{Quantitative comparison with the $\text{VP}^2$ baseline in the RoboDesk environment. }
    \label{fig5}
\end{figure*}

Since the predicted videos $V_t$ share the historical frame $O_t$, a lightweight visual tracker VT can be used to localize both the end-effector $e_\tau^n$ and the sub-goal $s_\tau^n$ in each future frame in all the action-conditioned predicted videos initialized with $e_t$, $s_t$, and $I_t$, formulated as:
\begin{equation}
\label{eq11}
\begin{split}
    \text{VT}(V_t | e_t, s_t, I_t) = \{&e_\tau^n, s_\tau^n, I_\tau^n | n\in \{1,2,...,N\}, \\ &\tau \in \{t+1, t+2, ..., t+T\}\}
\end{split}
\end{equation}
where we employ an efficient real-time tracking network SiamRPN~\cite{siamrpn} as the visual tracker in this work.

To encourage the robot to move towards the next sub-goal and avoid colliding with all the interference objects, we calculate the VLM-assisted cost $C_\text{VLM}^n$ as:
\begin{equation}\small
\label{eq12}
    C_\text{VLM}^n(t) = \sum\limits_{\tau=t+1}^{t+T}(||c(e_\tau^n) - c(s_\tau^n)||_2-||c(e_\tau^n)-c(I_\tau^n)||_2),
\end{equation}
\begin{equation}
\label{eq13}
C_\text{VLM}(t)=\{C_\text{VLM}^n(t)|n\in\{1,2,...,N\}\}
\end{equation}
where $c(\cdot)$ represents the center of the given bounding boxes.


 

\subsubsection{VLM Switcher}
The pixel distance cost can provide fine-grained guidance on the pixel level, and VLM-assisted cost fixes the gap in knowledge-level task planning. 
With these two sub-costs, we further design a VLM switcher with prompt $\phi_{D}$, which dynamically selects one or both appropriate sub-costs in each step $t$ adaptive to the current observation through knowledge reasoning to produce the final cost $C(t)$: 
\begin{equation}
\label{eq14}
    \text{VLM}(O_t, G\lor L|\phi_D) = w_D \in \{0,0.5,1\},
\end{equation}
\begin{equation}
\label{eq15}
    C(t) = w_D * C_P(t) + (1-w_D) * C_\text{VLM}(t)
\end{equation}

With the cost $C(t) = \{C^n(t)|n\in\{1,2,...,N\}\}$ as the assessment of all the action-conditioned videos, we select the candidate action sequence with the lowest cost for the following process. When the first action in this sequence is executed, the subsequent actions are fed into a global mean pooling layer to generate the sampling mean $\mu_t^{sub}$ to provide historical information in the action sampling of the next step.

Algorithm~\ref{algo} summarizes the whole process of the VLMPC framework. When the task is done or reaching the maximum time limit, the system will return an end signal.

\begin{figure}[t]
    \centering
    \includegraphics[width=1\linewidth]{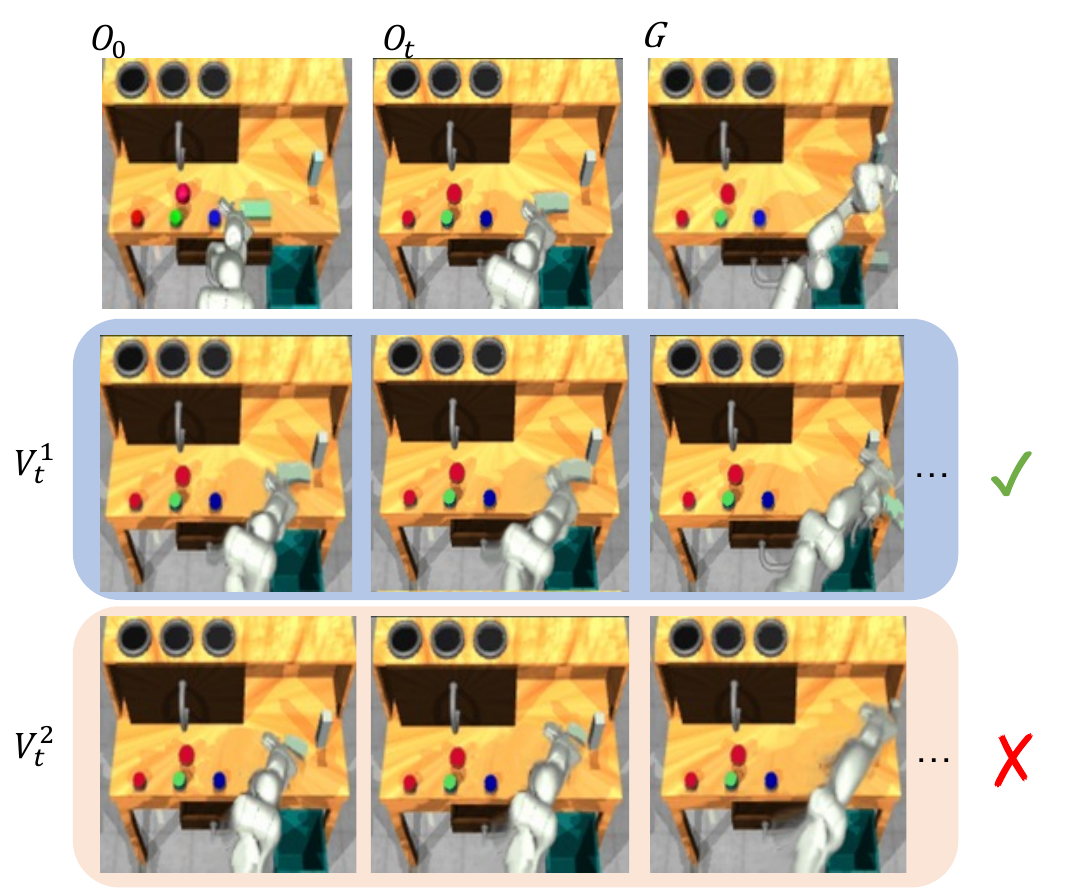}
    \caption{Visualization of the action-conditioned video predictions assessed by the hierarchical cost function of VLMPC via knowledge reasoning.}
    \label{fig6}
\end{figure}

\section{Experiments}

In this section, we first provide the implementation details of the proposed VLMPC framework. Then, we conduct two comparative experiments in simulated environments. The first is to compare VLMPC with VP$^2$ \cite{vp2} on 7 tasks in the RoboDesk environment \cite{kannan2021robodesk}. The second is to compare VLMPC with 5 existing methods in 50 simulated environments provided by the Language Table environment \cite{lynch2023interactive}. Next, we evaluate VLMPC in real-world scenarios. Finally, we investigate the effectiveness of each core component of VLMPC through ablation study. We provide the details of all the hyperparameters and the VLM prompts in the supplementary material.


\subsection{Implementation Details}

VLMPC employs Qwen-VL~\cite{qwen} and GPT-4V~\cite{2023GPT4VisionSC} as the VLMs. In the conditional action sampling module, VLMPC first uses GPT-4V to identify the target object with which the robot needs to interact, and then localizes the object through Qwen-VL. In the VLM-assisted cost, VLMPC first extracts the sub-goals and interference objects with GPT-4V, and then localizes them through Qwen-VL. The VLM switcher uses GPT-4V to dynamically select one or both sub-costs in each time step.

The training policy of the DMVFN-Act video prediction model contains 2 stages. In the first stage, we select 3 sub-datasets from the Open X-Embodiment Dataset~\cite{padalkar2023open}, a large-scale dataset containing more than 1 million robot trajectories collected from 22 robot embodiments. The 3 sub-datasets used for pre-training DMVFN-Act are Berkeley Autolab UR5, Columbia PushT Dataset, and ASU TableTop Manipulation. In the second stage, we collect 20 episodes of robot execution in the environment where the experiments are conducted and train DMVFN-Act to adapt to the specific scenario.

\subsection{Simulation Experiments}

\subsubsection{Simulation Environments and Experiment Settings}

The first evaluation is conducted on the popular simulation benchmark VP$^2$~\cite{vp2} which offers two environments RoboDesk~\cite{kannan2021robodesk} and \texttt{robosuite}~\cite{zhu2020robosuite}.
Considering the significant difference between the physical rendering of \texttt{robosuite} and real-world scenarios, we only use RoboDesk in this work. RoboDesk provides a physical environment with a Franka Panda robot arm, as well as 
a set of manipulation tasks. VP$^2$ conducts 7 sub-tasks: \emph{push \{red, green, blue\} button, open \{slide, drawer\}, push \{upright block, flat block\} off table.} For each sub-task, VP$^2$ provides 30 goal images as task input.

In the second experiment, we compare VLMPC with 5 existing methods in the Language Table environment~\cite{lynch2023interactive} on the \textit{move to area} task following VLP~\cite{du2024video}. Such a task is given by language instructions: \textit{move all blocks to different areas of the board}. The 5 competing methods are UniPi~\cite{du2023learning}, LAVA~\cite{lin2023videollava}, PALM-E~\cite{palm-e}, RT-2~\cite{rt-2} and VLP~\cite{du2024video}. We follow VLP \cite{du2024video} to compute rewards using the ground truth state of each block in the Language Table environment \cite{lynch2023interactive}. And we made the evaluation on 50 randomly initialized environments.

\subsubsection{Experimental Results}
The experimental results on the $\text{VP}^2$ benchmark are listed in Fig.~\ref{fig5}. It can be seen that VLMPC significantly outperforms the VP$^2$ baseline. We can see that for the tasks of \emph{push
\{red, green, blue\} button}, both the VP$^2$ baseline and VLMPC achieve high performance. This is simply because such tasks contain no multiple sub-goals. Thus, once the robot arm reaches the specific button and pushes it, the task is completed. On the other hand, the remaining tasks are more challenging, which require the robot to identify and move among multiple sub-goals as well as avoiding the collision with interference objects. We can see that VLMPC significantly outperforms the VP$^2$ baseline in such challenging tasks, demonstrating its good reasoning and planning capability.

\begin{table}[t]
\setlength\tabcolsep{1pt}
\linespread{1.2}
\begin{center}\small
\caption{Comparison with existing methods on the task of \textit{move to area}  in the Language Table environment.}

\begin{tabular}{c|cc}
\hline\hline
Method & Success Rate(\%)  & Reward \\
\hline\hline
{UniPi~\cite{du2023learning}} & 0 & 30.8 \\
{LAVA~\cite{lin2023videollava}} & 22 & 59.8 \\
{PALM-E~\cite{palm-e}} & 0 & 36.5 \\
{RT-2~\cite{rt-2}} & 0 & 18.5 \\
{VLP~\cite{du2024video}} & 64 & 87.3 \\
{VLMPC} & \textbf{70} & \textbf{89.3} \\

\hline\hline
\end{tabular}
\label{tab_baseline}
\end{center} 
\end{table}

\begin{figure*}[t]
    \centering
    \includegraphics[width=0.7\linewidth]{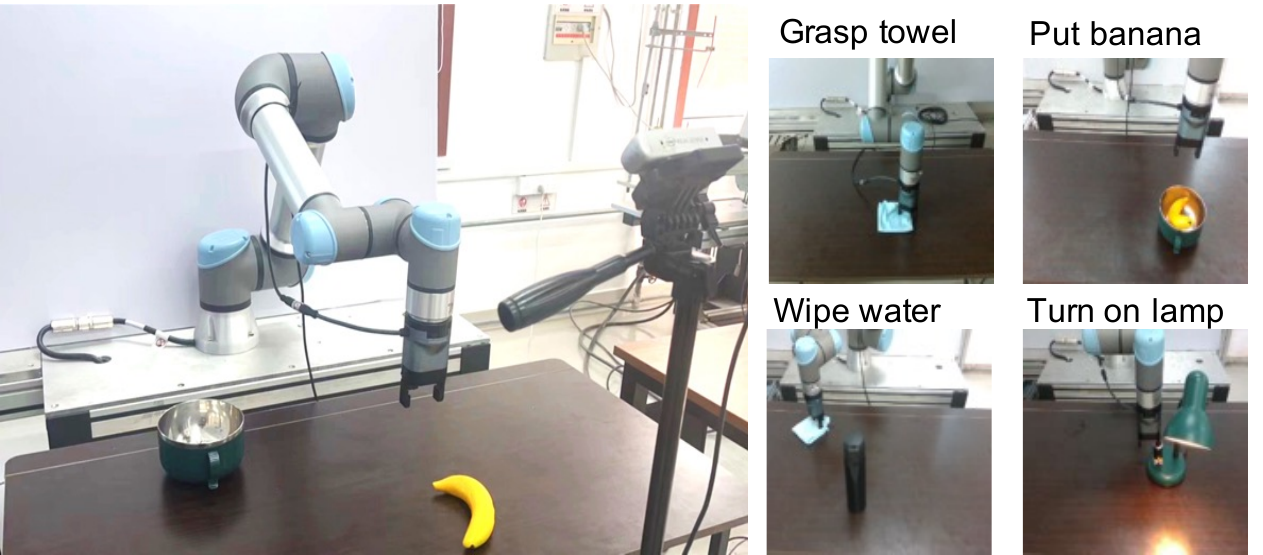}
    \caption{The real-world experimental platform includes a UR5 robot arm and a monocular RGB camera. It also shows a goal image for each of the four tasks.}
    \label{fig7}
\end{figure*}

Fig.~\ref{fig6} shows the visual results for the most challenging sub-task \emph{push flat}. This task requires pushing a flat green block off the table, while keeping other objects unmoved. We notice a slender block standing on the right edge of the table, which obviously serves an interference object. 
For the current observation $O_t$, we select two predicted videos for visualization. The second and the third rows of Fig.~\ref{fig6} show the predicted videos corresponding to different candidate action sequences. It can be seen that both candidate action sequences have the tendency to push the flat block off the table. It is noteworthy that the VP$^2$ baseline using a pixel-level cost and a simple state classifier assigns similar costs on both videos, which leads to the selection of an inappropriate action sequence. In contrast, VLMPC produces a higher cost for $V_t^2$ which contains a possible collision between the robot arm and the interference object. $V_t^1$ indicates a more reasonable moving direction and interaction with objects, and is thus assigned a lower cost. Such results demonstrate that the proposed hierarchical cost function can make desired assessment of the predicted videos on the knowledge level and facilitate VLMPC to select an appropriate action to execute. 

Table~\ref{tab_baseline} lists the quantitative results of the comparative experiment conducted in the Language Table environment~\cite{lynch2023interactive}, where the Reward metric is computed in accordance with the VLP reward \cite{du2024video}. It can be seen that the proposed VLMPC outperforms all competing methods. This is because VLMs are good at localizing specific areas. Therefore, through sampling actions towards the sub-goals, VLMPC enables the robot to successfully reach the sub-goals and complete the task.

\begin{figure*}[t]
    \centering
\includegraphics[height=10cm]{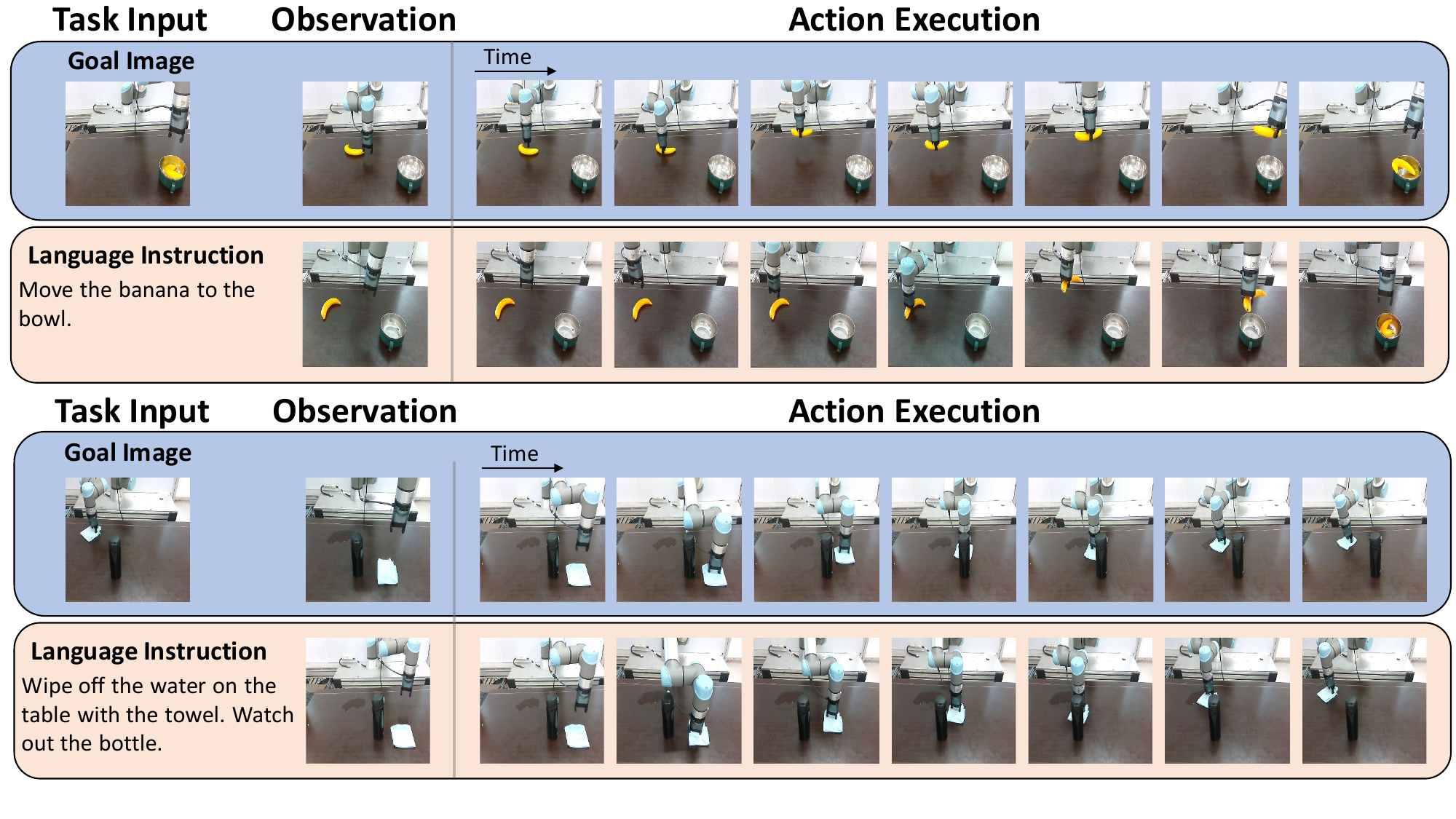}
    \caption{Visualization of the action execution for two challenging real-world manipulation tasks \emph{put the banana in the bowl} and \emph{wipe water}.}
    \label{fig8}
\end{figure*}

\subsection{Real-World Experiments}
\subsubsection{Experimental Setting}
As shown in Fig.~\ref{fig7}, we use a UR5 robot to conduct real-world experiments. A monocular RGB camera is set up in front of the manipulation platform to provide the observations. We design four manipulation tasks, including \emph{grasp towel}, \emph{put banana}, \emph{turn on lamp}, and \emph{wipe water}. It is noteworthy that the objects involved in these tasks are not included in the collected data for training the video prediction model.

In each manipulation task, the position of the objects is initialized randomly within the reachable space of the action, yielding different goal images. Fig.~\ref{fig7} shows some example goal images for the 4 tasks.



\begin{table}[t]
\setlength\tabcolsep{1pt}
\linespread{1.2}
\begin{center}\small
\caption{Results of VLMPC using goal image or language instruction as input in real-world experiments.}
\begin{tabular}{c|cc|cc}
\hline\hline
\multirow{2}{*}{Tasks} & \multicolumn{2}{c|}{Goal Image} & \multicolumn{2}{c}{Language Instruction} \\
 & Success Rate(\%)  & Time(s) & Success Rate(\%)  & Time(s) \\
\hline\hline
\emph{grasp towel} & 76.67 & 162.4 & 73.33 & 184.6 \\
\emph{put banana} & 60.00 & 203.9 & 46.67 & 230.7\\
\emph{turn on lamp} & 83.33 & 128.4 & 86.67 & 142.8\\
\emph{wipe water} & 36.67 & 289.3 & 23.33 & 331.9\\

\hline\hline
\end{tabular}
\label{tab1}
\end{center} 
\end{table}

\subsubsection{Experimental Results}
To properly evaluate VLMPC in real-world tasks, we repeat each task 30 times by randomly initializing the position of all objects and changes the color of the tablecloth every 10 times. We calculate the success rate and the average time for each task respectively. The results are listed in Table~\ref{tab1}. It can be seen that VLMPC achieves high success rates on the tasks of \emph{grasp towel} and \emph{turn on lamp}. These two tasks are relatively simple as there is no interference object in the scene. The success rates for the tasks of \emph{put banana} and \emph{wipe water} are low because they are more challenging. \emph{put banana} contains multiple sub-goals, and \emph{wipe water} is even more difficult as it involves both interference objects and multiple sub-goals. Such results demonstrate that VLMPC generalizes well to novel objects and scenes unseen in the training dataset.

We also provide the visual results for two challenging tasks \emph{put banana in the bowl} and \emph{wipe water}. As shown in Fig.~\ref{fig8}, in the \emph{put banana in the bowl} task, VLMPC correctly identifies the first sub-goal, \ie, the banana, based on the current observation, and drives the robot arm moving towards and finally grasping it. Then, VLMPC dynamically finds the next sub-goal, \ie, the bowl, and subsequently guides the robot to move to the area above it and opens the gripper. This example demonstrates VLMPC has the capability of dynamically identifying the sub-goals during the task.

The \emph{wipe water} task requires the robot arm to wipe off the water on the
table with the towel while watching
out the bottle. It is clear that this task contains two sub-goals \emph{towel} and \emph{water}, and an interference object \emph{bottle}. Fig.~\ref{fig8} shows that our method successfully identifies all of them, and guides the robot to select appropriate actions to execute while avoiding the collision with the interference object.

We provide more visualization results on four sub-tasks with both successful and failure cases, as well as related discussion in the supplementary material. We also provide video demonstrations in both simulated and real-world environments.

\subsection{Ablation studies}

We have conducted ablation study to demonstrate the effectiveness of each core component of VLMPC. In the experiments, we compare VLMPC with 4 variants described as follows:

\textbf{VLMPC-rs}: This is an ablated version of VLMPC where the conditional action sampling module is replaced with random sampling which simply sets the sampling mean $\mu_t$ to zero.

\textbf{VLMPC-PD}: This variant of VLMPC only uses the pixel distance cost as the cost function.

\textbf{VLMPC-VS}: This variant of VLMPC only uses the VLM-assisted cost as the cost function.

\textbf{VLMPC-MCVD}: In this variant of VLMPC, we replace DMVFN-Act with the action-conditioned video prediction model MCVD \cite{vp2,voleti2022mcvd}.

The results are shown in Table~\ref{tab_ablation}. First, compared with random sampling, our conditional action sampling module makes the robot complete various tasks more quickly and achieve higher success rates. This is because random sampling cannot make the sampled action sequences focus on the direction to sub-goals. Second, when VLMPC only uses the pixel distance cost, we found that the robot directly moves to the goal position and ignores intermediate sub-goals, leading to low success rates in the tasks \textit{put banana} and \textit{wipe water}. Besides, when VLMPC only uses the VLM-assisted cost, we found that VLM sometimes localizes incorrect sub-goals, which also leads to low success rates. Third, compared with DMVFN-Act, the diffusion-based video prediction model MCVD leads to much lower efficiency in all testing tasks.

\begin{table*}[t]
\setlength\tabcolsep{1pt}
\linespread{1.2}
\begin{center}\small
\caption{Ablation study using the variants of VLMPC on different tasks in real-world environments.}

\begin{tabular}{c|cc|cc|cc|cc}
\hline\hline
\multirow{2}{*}{VLMPC Variant} & \multicolumn{2}{c|}{\emph{grasp towel}} & \multicolumn{2}{c|}{\emph{put banana}} & \multicolumn{2}{c|}{\emph{turn on lamp}} & \multicolumn{2}{c}{\emph{wipe water}}\\
 & Success Rate(\%)  & Time(s) & Success Rate(\%)  & Time(s) & Success Rate(\%)  & Time(s) & Success Rate(\%)  & Time(s)\\
\hline\hline
{VLMPC-rs}  & 63.33 & 302.5 & 40 & 389.5 & 73.33 & 256.7 & 13.33 & 573.9\\
{VLMPC-PD} & 26.67 & 178.3 & 0 & - & 60.00 & \textbf{123.6} & 0 & -\\
{VLMPC-VS} & 56.67 & 201.5 & 46.67 & 297.3 & 56.67 & 243.7 & 10.00 & 543.9\\
{VLMPC-MCVD} & 33.33 & 509.3 & 23.33 & 689.4 & 46.67 & 553.8 & 6.67 & 803.5\\
{VLMPC} & \textbf{76.67} & \textbf{162.4} & \textbf{60.00} & \textbf{203.9} & \textbf{83.33} & 128.4 & \textbf{36.67} & \textbf{289.3}\\

\hline\hline
\end{tabular}
\label{tab_ablation}
\end{center} 
\end{table*}

\section{Conclusion} 
\label{sec:conclusion}
This paper introduces VLMPC that integrates VLM with MPC for robotic manipulation. It prompts VLM to produce a set of candidate action sequences conditioned on the knowledge reasoning of goal and observation, and then follows the MPC paradigm to select the optimal one from them. The hierarchical cost function based on VLM is also designed to provide an amenable assessment for the actions by estimating the future frames generated by a lightweight action-conditioned video prediction model. Experimental results demonstrate that VLMPC performs well in both simulated and real-world scenarios. 

A limitation of VLMPC lies in its process of video prediction, where a mismatch between the predicted video and the action sequence may occur and thereby affect the evaluation of the action sequences. Moreover, incorporating a large-scale VLM into each step of the MPC loop introduces higher computing cost inevitably. Without assessing the predicted video at each step, VLMPC cannot perfectly handle the tasks where the space of motion is strictly constrained. Hence, developing a more reliable video prediction model and designing a more efficient scheme for integrating VLM with MPC are of interest in the future work.

\section*{Acknowledgments}
This work was supported in part by the National Natural Science Foundation of China under Grants 61991411, U22A2057, and 62076148, in part by the National Science and Technology Major Project under Grant 2021ZD0112002, in part by the Shandong Excellent Young Scientists Fund Program (Overseas) under Grant 2022HWYQ-042, in part by the Young Taishan Scholars Program of Shandong Province No.tsqn201909029, and in part by Project for Self-Developed Innovation Team of Jinan City under Grant 2021GXRC038.

\bibliographystyle{plainnat}
\bibliography{references}

\end{document}